\documentclass[lettersize,journal]{IEEEtran}

\usepackage{amsmath,amsfonts}
\usepackage{amssymb}
\usepackage{algorithm}
\usepackage{algpseudocode}
\usepackage{array}
\usepackage[caption=false,font=normalsize,labelfont=sf,textfont=sf]{subfig}
\usepackage{textcomp}
\usepackage{stfloats}
\usepackage{url}
\usepackage{verbatim}
\usepackage{graphicx}
\usepackage{cite}
\usepackage{microtype}
\usepackage{booktabs} 
\usepackage{multirow}
\usepackage{makecell}
\usepackage{float}
\usepackage{pifont}
\usepackage{enumitem}
\usepackage{wrapfig}
\usepackage{xcolor}
\usepackage{textcomp}
\usepackage{bm}
\usepackage{mathtools}
\usepackage{dsfont}  
\usepackage{bbm}     
\usepackage{hyperref}
\usepackage{balance}

\newcommand{\ours}{\textsc{ICAFS}}

\newcommand{\bx}{\mathbf{x}}
\newcommand{\bz}{\mathbf{z}}

\newcommand{\ts}{\tilde{s}}
\newcommand{\ty}{\tilde{y}}

\newcommand{\cS}{\mathcal{S}}

\begin{document}

\title{ICAFS: Inter-Client-Aware Feature Selection for Vertical Federated Learning}

\author{Ruochen Jin\textsuperscript{1,2}, 
    Boning Tong\textsuperscript{1}, 
    Shu Yang\textsuperscript{1}, 
    Bojian Hou\textsuperscript{1,*}, 
    Li Shen\textsuperscript{1,*},\\
    for the Alzheimer’s Disease Neuroimaging Initiative\textsuperscript{3}
        
\thanks{\textsuperscript{1}University of Pennsylvania, Philadelphia, PA, USA.

\textsuperscript{2}East China Normal University, Shanghai, China.

\textsuperscript{3}Data used in preparation of this article were obtained from the Alzheimer’s  Disease Neuroimaging Initiative (ADNI) database (adni.loni.usc.edu). As such, the investigators within the ADNI  contributed to the design and implementation of  ADNI and/or provided data but did not participate in  analysis or writing of this report.  A complete listing  of ADNI investigators can be found at: \url{http://adni.loni.usc.edu/wp-content/uploads/how_to_apply/ADNI_Acknowledgement_List.pdf}

\textsuperscript{*}Corresponding authors:

bojian.hou@pennmedicine.upenn.edu,  
li.shen@pennmedicine.upenn.edu.
}
}

\markboth{Journal of \LaTeX\ Class Files,~Vol.~14, No.~8, August~2021}%
{Shell \MakeLowercase{\textit{et al.}}: A Sample Article Using IEEEtran.cls for IEEE Journals}


\maketitle

\begin{abstract}
Vertical federated learning (VFL) enables a paradigm for vertically partitioned data across clients to collaboratively train machine learning models. 
Feature selection (FS) plays a crucial role in Vertical Federated Learning (VFL) due to the unique nature that data are distributed across multiple clients.
In VFL, different clients possess distinct subsets of features for overlapping data samples, making the process of identifying and selecting the most relevant features a complex yet essential task.
Previous FS efforts have primarily revolved around intra-client feature selection, overlooking vital feature interaction across clients, leading to subpar model outcomes.
We introduce \ours{}, a novel multi-stage ensemble approach for effective FS in VFL by considering inter-client interactions.
By employing conditional feature synthesis alongside multiple learnable feature selectors, \ours{} facilitates ensemble FS over these selectors using synthetic embeddings. This method bypasses the limitations of private gradient sharing and allows for model training using real data with refined embeddings.
Experiments on multiple real-world datasets demonstrate that \ours{} surpasses current state-of-the-art methods in prediction accuracy. 
\end{abstract}

\section{Introduction}

Federated learning (FL), a paradigm that enables distributed machine learning without revealing local private data, has found extensive applications across various domains \cite{li2020review,qin2021federated,ahmed2021federated}. A specific form of FL, known as vertical federated learning (VFL) \cite{chen2020}, deals with scenarios where different entities possess data with common samples but distinct feature spaces. This approach is particularly vital in the fields that depend on the analysis of high-dimensional data, such as medicine and biology \cite{jackson2018genetic}, where it has become increasingly prevalent. An illustrative example of VFL is in collaborative medical research, where some medical centers contribute genomics data and others provide imaging data for the same patient cohort, thereby enhancing the disease diagnosis \cite{Wu2023FederatedAL}.

In machine learning practice, the identification of target-related features is crucial in many applications, such as biomarker discovery for the design of clinical treatments \cite{jackson2018genetic}. The technique that discovers features responsible for the target variable is known as feature selection (FS) \cite{jiang2017high,li2015feature,zhang2018feature,sun2019high}.
Existing studies focused on the centralized setting, where algorithms have full access to the entire dataset, allowing for comprehensive analysis and selection of the most relevant features for the targeted task. 
These FS techniques can be broadly classified into three distinct categories: 1) filter-based methods, which evaluate individual feature relevance using statistical metrics such as Gini impurity, and filter out irrelevant features before model training \cite{chen2017kernel,song2012feature,estevez2009normalized}; 2) wrapper-based methods, which explore extensive search spaces to identify the optimal subset of features \cite{roy2015feature,kabir2010new}; and 3) embedding-based methods, which aim to simultaneously select essential features and learn the model, integrating the feature selection process within the model training \cite{yamada2020feature,li2016deep,hans2009bayesian}.
There is a growing interest in adapting these methods for decentralized settings. This shift presents unique challenges in maintaining data privacy while ensuring effective feature analysis.

Filter-based feature selection methods offer several key advantages over embedding-based or wrapper-based approaches, making them a compelling choice for many applications. These methods are computationally efficient, model-independent, and highly scalable, allowing them to handle high-dimensional datasets with ease. Their simplicity makes them easy to implement and interpret, while their independence from specific learning algorithms helps prevent overfitting. Importantly, filter methods preserve the original feature space, maintaining feature interpretability and allowing for direct validation by domain experts. This preservation of the original features is particularly valuable in fields like healthcare and scientific research, where understanding the specific features driving predictions is crucial. By offering these benefits, filter-based feature selection methods provide a robust and interpretable approach to identifying relevant features, especially in scenarios with limited labeled data or high-dimensional datasets \cite{jiang2017high,li2015feature,zhang2018feature,sun2019high}.

Facing the increased feature dimensionality in VFL compared with learning within a single client, training a global model in VFL is more vulnerable to noisy and spurious features. 
Recent studies have explored FS in VFL settings~\cite{feng2020multi,chen2021fedeini,hou2022,zhang2022a,zhang2022b,chen2022,li2023,feng2022}. These methods determine the important features based on their local model performance within each client only. However, neglecting the interactions between clients' features will restrict the overall performance of the FS. For example, on the one hand, an unselected feature determined by client A's local model can be important when considering client B's feature together for the global model; on the other hand, some of the selected features from client A and B can be highly correlated, resulting in redundancy. Furthermore, the correlation among feature sets
from different clients
will become more complicated as the number of clients increases. Therefore, FS should be performed considering the inter-client correlations. 

To address the limitation of previous methods that only focused on individual clients, a recent method named LESS-VFL has investigated the possibility of concurrently learning FS across multiple clients in the embedding space~\cite{castiglia2023}. 
However, despite its superior prediction performance,
this method requires sharing of raw gradients (\textit{i.e.}, the gradient of loss on the true label with respect to model parameters and data features) across all the clients. This may put clients at risk of data leakage, making them vulnerable to attacks like label inference~\cite{fu2022label}, gradient inversion~\cite{liu2022batch}, data reconstruction~\cite{vu2023}, and other similar threats. 
Therefore, designing a suitable FS strategy that does not require raw gradient sharing is crucial but difficult in the field of VFL.

\begin{figure}
    \centering
    \includegraphics[width=\columnwidth]{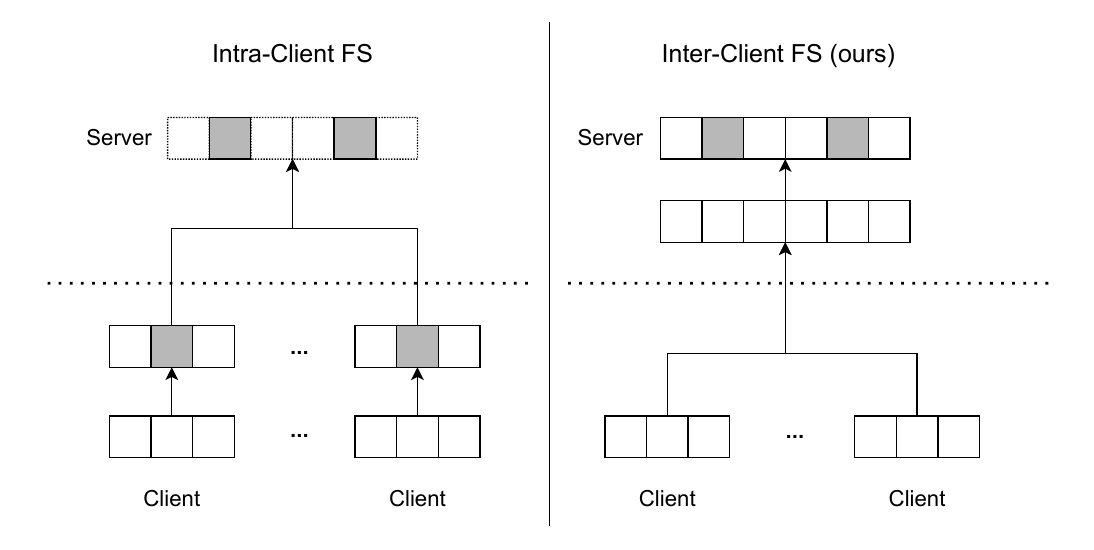}
    \caption{VFL feature selection architecture comparison between Intra-Client FS and Inter-Client FS (ours).
    Significant embedding components (in gray) are selected.
    In the left panel, each client selects their own features. In the right panel, the selection is processed after concatenating embeddings from clients.
    }
    \label{fig example}
\end{figure}

In this work, we consider the inter-client FS setting in VFL where the concatenated embeddings from clients can be viewed as the inputs for the global server model and perform FS interactively based on server model outputs (Fig. \ref{fig example} right).
The inter-client mechanism might present a greater challenge to preserving privacy due to increased data sharing.
We seek to solve the challenging problem \emph{``How to perform FS in VFL that considers inter-client feature interaction while avoiding private information sharing for training neural networks?"} 
To this end, we propose Inter-Client-Aware Feature Selection, referred to as \ours{} which includes three stages. 
In Stage 1 (Sec.~\ref{sec:syn}), to avoid asking the client to share raw gradients with others, we use federated global conditional feature synthesis~\cite{zhao2023} to generate synthetic data. Note that there is no private gradient sharing in this stage. In Stage 2 (Sec.~\ref{sec:emb_slct}), we propose to use filter-based FS method to do feature selection. Unlike LESS-VFL~\cite{castiglia2023} or other shrinkage or embedding-based FS methods,
the filter-based FS can avoid overfitting when working on small data. This approach involves introducing multiple learnable gates\footnote{The feature selector is essentially parameterized by the learnable gate.} that are trained using the synthetic features and labels for FS backpropagation. The automatic gating mechanism takes into account the correlation structure of input features, enhancing the model's decision-making process. The ensemble of different feature selectors further guarantee the robustness of the final prediction. In Stage 3 (Sec.~\ref{sec:classification}), we use the feature selectors learned from Stage 2 to select features from real data. They are further used to refine the prediction heads in the server and the local encoders, eliminating the need for private gradient sharing.

Our method \ours{} exhibits several distinctive and superior characteristics when compared to other state-of-the-art FS in VFL algorithms listed in Table~\ref{tab:comparison}.  
\begin{itemize}
    \item \textbf{Firstly}, it is feasible to apply to neural networks, making it a versatile tool in commonly used VFL settings with deep learning algorithms.
    \item \textbf{Secondly}, it is designed to be task-aware, improving the relevance and performance of selected features for the given task.
    \item \textbf{Thirdly}, it performs inter-client feature selection, leading to a more comprehensive feature set that leverages the strength of distributed data sources.
    \item \textbf{Lastly}, it operates without the need for private gradients, which is facilitated via our novel federated conditional feature synthesis. 
\end{itemize}
While individual elements draw inspiration from existing techniques, their integration creates a unique approach that balances feature selection accuracy, privacy, and efficiency. This combination allows ICAFS to overcome the limitations of previous methods, particularly in handling complex neural network architectures and preserving privacy in multi-client scenarios. Comprehensive experiments on real-world datasets demonstrate the effectiveness of this integrated approach across various settings.

\begin{table}
\centering
\caption{VFL feature selection algorithms.}
\label{tab:comparison}
\resizebox{\columnwidth}{!}{
\begin{tabular}{l|cccc}
\toprule
\makecell{VFL FS \\ Algorithm} & \makecell{Neural \\networks} & \makecell{Task-aware \\ (in-training) }&\makecell{Inter-client \\ FS} &\makecell{w/o private \\ gradient} \\ \midrule
MMVFL(2020)                    &         \color{red}\ding{53}                 &       \color{blue}\checkmark                  &       \color{red}\ding{53} & \color{blue}\color{blue}\checkmark \\
Fed-EINI(2021)                     &         \color{red}\ding{53}                 &       \color{blue}\checkmark                  &       \color{red}\ding{53} & \color{blue}\checkmark \\
Hou et al.(2022)                     &         \color{red}\ding{53}                 &       \color{blue}\checkmark                  &       \color{red}\ding{53}   & --                                    \\
Zhang et al.(2022a)                   &         \color{red}\ding{53}                 &       \color{blue}\checkmark                  &       \color{red}\ding{53}      &  \color{red}\ding{53}                                 \\
Zhang et al.(2022b)                  &         \color{blue}\checkmark       &         \color{red}\ding{53}                          &           \color{red}\ding{53}        &        \color{blue}\checkmark                     \\
EVFL(2022)                       &         \color{blue}\checkmark       &        \color{red}\ding{53}                           &         \color{red}\ding{53}         &        \color{blue}\checkmark                                \\
VFLFS(2022)                   &         \color{blue}\checkmark       &       \color{blue}\checkmark                  &             \color{red}\ding{53}                & \color{blue}\checkmark                \\
FedSDG-FS(2023)                 &         \color{blue}\checkmark       &  \color{red}\ding{53}  &  \color{red}\ding{53}                  &             \color{blue}\checkmark                                  \\ 
LESS-VFL(2023) & \color{blue}\checkmark & \color{blue}\checkmark & \color{blue}\checkmark  & \color{red}\ding{53}  \\ \midrule
\ours (ours)                     &         \color{blue}\checkmark       &       \color{blue}\checkmark                  &             \color{blue}\checkmark   &  \color{blue}\checkmark                     \\ \bottomrule
\end{tabular}}
\end{table}

\section{Related Work}
\noindent\textbf{Different capabilities of Current Feature Selection in Vertical Federated Learning:}
VFL has emerged as a powerful paradigm for distributed learning across multiple clients without sharing raw data. However, the current state-of-the-art methods for feature selection in VFL have several drawbacks, particularly concerning the handling of correlations from different clients, see Table \ref{tab:comparison}. There is a growing interest in designing FS methods for training neural networks in VFL. It is important to take into account the interaction of features between clients and to maintain data privacy, which means that private gradients should not be shared across clients. However, none of the existing methods meet these requirements.

    \textbf{Learnable Selection:} In the field of VFL, several significant contributions have been made, but most of them lack exploration
    for neural network-based models.
    Feng and Yu \cite{feng2020multi} introduced a multi-participant multi-class VFL framework. Their approach enables collaboration between multiple parties with different feature spaces but does not extend to neural network architectures, limiting its applicability in deep learning scenarios. Chen et al. \cite{chen2021fedeini} proposed Fed-EINI, an efficient and interpretable inference framework for decision tree ensembles in VFL. While the method offers interpretability, it is specifically tailored to decision trees and does not provide a mechanism to incorporate neural networks. Hou et al. \cite{hou2022} developed a verifiable privacy-preserving scheme based on vertical federated random forest. Their approach ensures privacy in federated learning but is confined to random forest models, excluding the potential benefits of neural network-based learning. Zhang et al. \cite{zhang2022a} presented an embedded vertical-federated feature selection algorithm based on particle swarm optimization. Although innovative in feature selection, the method does not consider neural network models, restricting its utility in modern deep learning applications.
    
     \textbf{Features selected during training:} Several works have been conducted in the domain of vertical federated learning (VFL), with a focus on feature selection. However, a common limitation among these works is that feature selection is not integrated into the training process, which can lead to suboptimal performance. Chen et al. \cite{chen2022} introduced EVFL, an explainable VFL for data-oriented artificial intelligence systems. Their method provides explainability but lacks an integrated feature selection mechanism during training, which might lead to inefficiencies in handling high-dimensional data. Li et al. \cite{li2023} presented FedSDG-FS, an efficient and secure feature selection approach for VFL. Although their method is designed for efficiency and security, the separation of feature selection from the training process may not fully leverage the interdependencies between features and the target model, especially in complex neural network architectures.


\textbf{Gate Mechanism in Feature Selection:} The gate mechanism has recently been explored as a novel approach to feature selection, which is a filter-based selection with learnable parameters and pre-fixed threshold. Lee et al.~\cite{lee2021} introduced a self-supervision enhanced feature selection method with correlated gates, providing a new perspective on handling correlations in feature selection. Imrie et al.~\cite{imrie2022} further extended the concept of gate mechanisms in composite feature selection using deep ensembles, with the gate mechanism a learnable selection during training can be achieved. Compared with conventional selection algorithms, e.g. lasso, gate mechanism is more effective and robust.

\textbf{Correlation Handling:} 
Current methods frequently neglect to consider the interdependencies among features held by different clients. For example, the work of Zhang et al.~\cite{zhang2022b} prioritized security yet failed to account for these feature correlations during selection. Up until now, LESS-VFL \cite{castiglia2023} is recognized as the initial algorithm to implement LASSO in embedding components, addressing correlations across various clients. However, lasso, being a shrinkage-embedded feature selection technique, necessitates a substantial parameter count, which can lead to overfitting, particularly in scenarios with high-dimensional data or a constrained quantity of training samples. Our work differs from these methods by considering the challenge of both correlations across various clients and privacy leakage.

\begin{table}[t]
\centering

\caption{Summary of Notation
}
\begin{tabular}{ll}
\hline
Notation                                                            & Definitions                             \\ \hline
$K$ & Number of clients.                      \\
$N$                                                                  & Number of learners.                     \\
$f(\cdot)$                                             & Encoder model.                          \\
$h(\cdot)  $                                           & Prediction model.                       \\
$D$                                                                   & Discriminator model.                    \\
$G$                                                                   & Generator model.                        \\
$\sigma(\cdot)$                                              & Sigmoid function.        \\
$\Theta$                                               & The set of all learnable parameters. \\
$\textbf{x}$, $\tilde{\textbf{x}}$  & Real and synthetic local embedding. \\
$\textbf{z}$, $\tilde{\textbf{z}}$                                                                  &  Real and synthetic concatenation embedding.         
\\
$\textbf{s}$, $\tilde{\textbf{s}}$  & Selected real and synthetic features. \\
$y$, $\tilde{y}$, $\hat{y}$  & Real label, synthetic label, and ensemble prediction. \\
$\bm\alpha$  & Learnable gate. \\
$\textbf{m}$  &  Binary selection vector. \\
CV  &  Conditional vector. \\
$\epsilon$ & Noise used for generating synthetic data. \\
$\mathcal{D}$, $\tilde{\mathcal{D}}$ & Real and synthetic dataset. 
\\ \hline
\end{tabular}
\end{table}

\section{Problem Formulation and Background}

In this section, we introduce the concept of Feature Selection (FS) within the context of Vertical Federated Learning (VFL), focusing on its application in the embedding space of neural networks.

\textbf{Vertical Federated Learning (VFL).}
Vertical Federated Learning (VFL) differs from Horizontal Federated Learning (HFL or simply FL) in that it divides data based on features rather than samples. In a VFL system, there are \( K \) clients, each possessing \( N \) samples that are aligned across clients through shared sample IDs. For the \( k \)-th client, the input feature dimension is denoted as \( d_k \), and after feature embedding, the output feature dimension is \( d^\prime_k \). The input features for client \( k \) are represented by \( \bx_k = [x_{k}^1, \dots, x_{k}^{d_k}]^\top \in \mathbb{R}^{d_k} \).

Among these clients, only one, referred to as the \textit{target client}, possesses the labels for the samples, denoted as \( y \in \mathcal{Y} \). The target client's dataset is thus denoted as \( \mathcal{D}_c = \{(\bx_{c,i}, y_i)\}_{i=1}^{N} \), while the other clients have datasets containing only the input features, denoted as \( \mathcal{D}_{\bar{c}} = \{\bx_{\bar{c},i}\}_{i=1}^{N} \) for \( \bar{c} \in [K] \setminus c \). The complete data available for VFL training is \( \mathcal{D} \triangleq \{\mathcal{D}_k\}_{k=1}^{K} \).

The \textit{primary goals} of VFL are twofold: (1) to utilize the features from all clients to enhance the prediction accuracy of the target client's labels, and (2) to preserve data privacy and model confidentiality of all clients. In a general VFL framework, each client constructs a local feature extractor \( f_{\theta_k}:\mathbb{R}^{d_k} \mapsto \mathbb{R}^{d^\prime_k} \), parameterized by \( \theta_k,k\in[K] \). The output of this feature extractor for the \(i\)-th sample, \( \bz_{k,i} = f_{\theta_k}(\bx_{k,i}) \in \mathbb{R}^{d^\prime_k} \), represents the extracted features. These extracted features from all clients are concatenated into a single vector \( \bz_i = [\bz_{1,i}^\top; \dots; \bz_{K,i}^\top]^\top\in\mathbb{R}^{p} \), where \( p = \sum_{k=1}^{K} d_k^\prime \), which serves as the input to the server's fusion model \( h_{\theta_s}: \mathbb{R}^{p} \mapsto \mathcal{Y} \). The server model, parameterized by \( \theta_s \), predicts the label for each sample.
Although VFL can leverage the information from all the clients to facilitate prediction, it usually neglects the redundant feature information across the clients. Therefore, it is beneficial to conduct feature selection during the VFL procedure. In our framework, we perform feature selection (FS) on the extracted features.
This approach enhances efficiency by reducing dimensionality and computational complexity, improves privacy by working with transformed data, and potentially boosts model performance by focusing on more meaningful patterns.
 The concatenated feature vector for the \(i\)-th sample is represented as \( \bz_i = \texttt{concat}(\{\bz_{k,i}\}_{k=1}^{K}) = [\bz_{1,i}^\top; \dots; \bz_{K,i}^\top]^\top \in \mathbb{R}^p \). The objective of FS is to identify a subset of these features, \( \mathcal{S} \subset \{1,\ldots,p\} \), that are most relevant for predicting the target label. This is achieved through a selection function \( \mathcal{F} \), which operates using a binary selection vector \( \textbf{m}_k \in \{0,1\}^{d_k^\prime} \) for each client \(k\).

The global selection function can be expressed as:
\begin{align}
    \mathcal{F}([\bz_{1,i}^\top; \dots; \bz_{K,i}^\top]^\top) \triangleq [\textbf{m}_1 \odot \bz_{1,i}^\top, \dots, \textbf{m}_K \odot \bz_{K,i}^\top]^\top,
    \label{eq:gate}
\end{align}
where \( \odot \) denotes the Hadamard (element-wise) product. The selection vectors \(\{\textbf{m}_k\}_{k=1}^K\) are learned jointly with the global model. The selected features are then used for prediction through \( \hat{y}_i = h_{\theta_s}(\mathcal{F}(\bz_i)) \). The learnable parameters \( \Theta = \{\{\theta_s^n\}_{n=1}^N, \{\theta_k\}_{k=1}^{K}, \{w^n\}_{n=1}^N\} \) are optimized by minimizing the loss \( \ell(y_i, \hat{y}_i) \), which measures the discrepancy between the predicted label \( \hat{y}_i \) and the true label \( y_i \).

The design of the selection function and the associated training process are critical components of the proposed method, which will be discussed in detail in Section~\ref{sec:emb_slct}. 

\begin{figure*}
    \centering
    \includegraphics[width=1\linewidth]{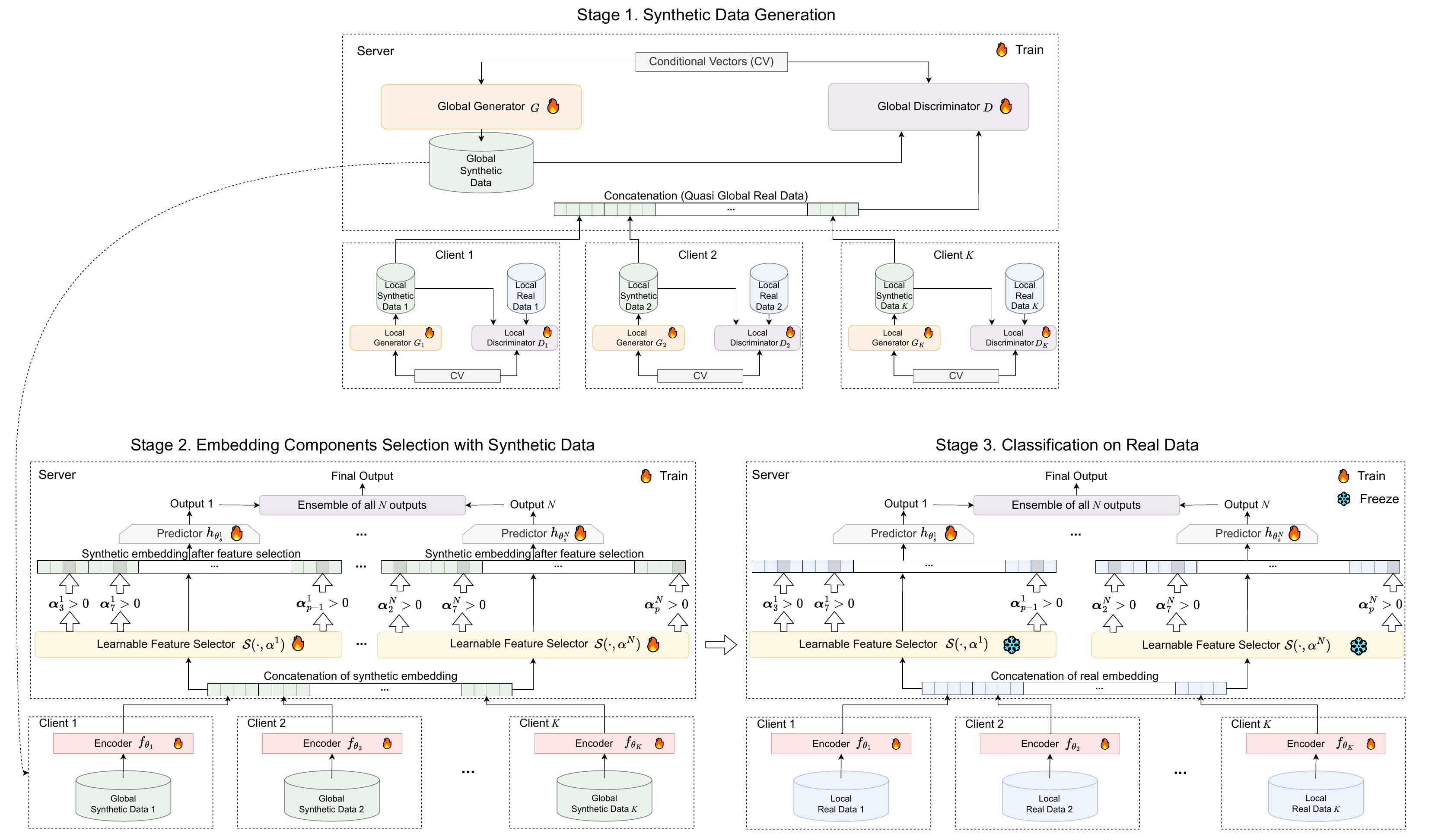}
    \caption{Illustration of \ours~pipeline. There are three stages including (1) synthetic data geneartion, (2) embedding components selection with synthetic data, and (3) classification on real data. Significant embedding components (emphasized by gray with $\bm\alpha>0$) are selected after training. The server-side and client-side models are represented up and down, respectively. Synthetic data generated in Stage 1 will be used in Stage 2.}
    \label{fig:framework}
\end{figure*}

\section{Method}
The overall pipeline of \ours~contains three stages, including \textit{1) Synthetic data generation:} synthetic data and labels are generated globally by a pre-trained Wasserstein GAN. \textit{2) Embedding components selection:} embedding components are calculated from synthetic data and afterward sent to the server. The server concatenates embedding components, performs selection with a gate filter, and updates gate vectors from synthetic data. Meanwhile, we fixed the gate filter and performed a selection of real embeddings. \textit{3) Classification on real data:} predictor is updated by real data, and the server sends the prediction back to each client, and then each client updates its own encoder locally. We depict the overview of \ours~pipeline in Figure \ref{fig:framework}. Notably, we consider a targeted client as the server in our setting. Therefore, there is no need to share any labels or real logits.

\subsection{Synthetic Data Generation}\label{sec:syn}
This stage involves generating synthetic data and labels globally using a pre-trained model. The aim is to enable collaborative machine learning across clients while maintaining data privacy. By using synthetic data, the real data’s private gradients remain local, and only the synthetic data's gradients are exchanged between clients and the server during training and classification.


Inspired by the works of Wu et al. \cite{wu2022} and Zhao et al. \cite{zhao2023}, we employ a Wasserstein GAN with gradient penalty and conditional vectors (CVs). The continuous variables are modeled using a variational Gaussian mixture model, while categorical variables are one-hot encoded. The CVs are formulated to identify continuous and discrete features.

The synthetic data generation process in each iteration involves two steps:
\begin{enumerate}
    \item Each client $k$ generates its synthetic data from random Guassian noise $\xi$ and CV: \(\tilde{\bx}_k = G_k(\epsilon,\text{CV})\) using their local generator \(G_k\), and compares it to their real data \(\bx_k\) using a local discriminator \(D_k\). The discriminator loss \(\mathcal{L}_{\rm D}(D_k(\tilde{\bx}_k), D_k(\bx_k))=\mathbb{E}_{\mathcal{D}}[ D_k(\bx_k)-D_k(\tilde{\bx}_k)]\) is maximized to improve the authenticity of the synthetic data. The generator loss \(\mathcal{L}_{\rm G}(D_k(\tilde{\bx}_k))=\mathbb{E}_{\mathcal{D}}[D_k(\tilde{\bx}_k)]\) is used to minimize the log probability of the discriminator correctly identifying the synthetic data as fake.
    
    \item The server concatenates the synthetic data from all clients \(\tilde{\bx}_s = [\tilde{\bx}_1, \dots, \tilde{\bx}_K,\tilde{y}]\) as the quasi global real data and compares it with the global synthetic data generated by the server \(G_s(\xi, \text{CV})\) using a global discriminator \(D_s(\tilde{\bx}_s)\). The discriminator loss and generator loss here are calculated in the same way as the local corresponding losses shown in 1) above.
\end{enumerate}
The update of the local discriminators and generators in Step 1 and the update of the global discriminators and generators in Step 2 are conducted iteratively to improve the quality of the synthetic data. The synthetic data generation process is summarized in Algorithm~\ref{alg:vfl_gan}.

\begin{algorithm}[t]
\small
\caption{\small ICAFS Stage 1: Synthetic Data Generation}
\label{alg:vfl_gan}
\begin{algorithmic}[1]
\State \textbf{Input: }  
Real data ${\mathcal{D}}=(\{{X}_k\}_{k=1}^K, {Y})$, random Guassian noise $\xi$ and conditional vector CV.
\State \textbf{Output:}
Synthetic data $\tilde{\mathcal{D}}=(\{\tilde{X}_k\}_{k=1}^K, \tilde{Y})$.

\State Initialize all network parameters
\For{iteration $t = 1, \dots,T$}

    \State // \textit{Step 1: Client-side Local Training}
    \For{client $k = 1, \dots,K$ \textbf{in parallel}}
        \State Generate synthetic data: $\tilde{\bx}_k \leftarrow G_k(\xi, \text{CV})$
        
        \State Update local discriminator $D_k$ via
        maximizing discriminator loss: 
            $\mathcal{L}_{D_k} = \mathbb{E}[D_k(\bx_k)] - \mathbb{E}[D_k(\tilde{\bx}_k)]$
        
        \State Update local generator $G_k$ via minimizing generator loss: $\mathcal{L}_{G_k} = \mathbb{E}[D_k(\tilde{\bx}_k)]$
    \EndFor
    
    \State // \textit{Step 2: Server-side Global Training}
    \State Aggregate synthetic data: $\tilde{\bx}_s \leftarrow [\tilde{\bx}_1, \dots, \tilde{\bx}_K, \tilde{y}]$
    \State Generate global synthetic data: $\tilde{\bx}_g \leftarrow G_s(\xi, \text{CV})$
    
    \State Update global discriminator $D_s$ via
maximizing loss $\mathcal{L}_{D_s} = \mathbb{E}[D_s(\tilde{\bx}_s)] - \mathbb{E}[D_s(\tilde{\bx}_g)]$
    
    \State Update global generator $G_s$ via
minimizing loss $\mathcal{L}_{G_s} = \mathbb{E}[D_s(\tilde{\bx}_g)]$
\EndFor
\State Generate synthetic data $\tilde{\mathcal{D}}$ using $G_s$
\end{algorithmic}
\end{algorithm}

\subsection{Embedding Components Selection with Synthetic Data}\label{sec:emb_slct}
This stage begins with local representation learning, where each client uses an encoder to extract features from their data. The goal is to share useful information among clients while preserving privacy.

Embedding component selection from synthetic data is crucial for \ours. This approach takes into account both inter-client feature interactions and prevents the sharing of private gradients. However, identifying the combination of features that is most useful for the targeted task is an NP-hard problem~\cite{amaldi1998approximability}. For example, given $d$ features, there are $2^d-2$ ways of selecting a subset of the features, which is particularly challenging in VFL where $d$ can be a large number when the number of clients increases. To tackle the complexity challenge, we borrow the strategy from CompFS~\cite{imrie2022} that automatically learns different gates and ensembles them for final prediction. We employ this strategy to be integrated into our server module, given it does not require prior knowledge of the feature and label pairs, making it more flexible and privacy-preserving for VFL applications. 
Next, we detail the embedding component selection procedure.

The process of embedding component selection involves the following steps:
\begin{enumerate}
    \item Gating process of multiple selectors:
    In order to explore the various potential combinations, we create multiple selectors to select different sets of features. For this purpose, we propose to use $N$ feature selectors $\cS^n(\cdot; \bm\alpha^n),\ i=1,\ldots,N$ with $N$ learnable gates $\{\bm{\alpha}^n\}_{n=1}^N$ (we will introduce how to learn these gates in next step). Consequently, we have $N$ embeddings $\{\tilde{s}^n\}_{n=1}^N$ generated:
    \begin{align} 
    \ts^n = \cS^n(\tilde{\textbf{z}}; \bm\alpha^n) = {\bm\alpha}^n \odot \tilde{\textbf{z}}+ (1-{\bm\alpha}^n) \odot \bar{\textbf{z}}.
    \label{selection}
    \end{align}
    In this equation, every element in $\bar{\textbf{z}}$ is identical, representing the average value across all $p$ dimensions. Each component in $\tilde {\textbf{z}}$ will be replaced by a weighted average of the original value and the mean value across all $p$ dimensions. When the weight $\alpha_i$ for the $i$th component is low or nearly 0, the resulting impact for this component can be seen as being replaced by the mean value, which indicates that this feature is not selected. The reason we use the mean value instead of zero value for the replacement is that zero can possess certain semantic meanings in certain tasks. For example, zero value of Celsius degree indicates the freezing point. 
    \item Learnable Gate for Embedding Selection: We introduce learnable gates \(\bm\alpha^n \in [0,1]^p,n=1,\ldots,N\) to efficiently optimize feature selection. The local encoder \(f_{\theta_k}\) processes synthetic input \(\tilde{x}_k\) to produce embedding \(\tilde{z}_k = f_{\theta_k}(\tilde{x}_k)\). The server concatenates these embeddings and converts them into low-dimensional vectors. The values in each gate \(\bm\alpha\) are continuous to guarantee receiving valid gradients during training: 
    \begin{align}\label{eq:alpha}
    \alpha^n_i =  {\sigma(\tau w^n_i z_i)}/{\sigma( w_i^{(0)})}, \quad \tau = \gamma^{t/T} \ {\text{and}} \  i \in [p],
    \end{align}
    where $\textbf{w} = [w_1, \dots, w_p]^\top \in \mathbb{R}^{p}$
    is the learnable parameter, $\sigma(\cdot)$ represents the sigmoid function applied element-wisely to approximate a binary gate, $t$ is the current training epoch number, $T$ is the total training epoch and $\gamma$ is a hyperparameter to control the temperature $\tau$ and $\tau$ will turn to $\gamma$ after training for $T$ epochs. As the training proceeds, $\tau$ will become larger and larger, which will make the training procedure more stable \cite{lyu2023}.
    This would allow the continuous gating vector $\bm\alpha$ to receive valid gradients in early stages, yet increasingly approximate binary gate as the epoch number $t$ grows. 
    \item Aggregation of selectors: The server aggregates the selected embeddings from all selectors using a linear prediction mechanism and element-wise summation to form the final prediction. The ensemble’s prediction is a combination of the outputs of all selectors.
    Consequently, the ensemble's prediction can be expressed as:
\begin{align}
    \hat{y} = \rho\left[\sum_{n=1}^{N} h_{\theta^n_s}(\tilde{s}^n)\right] 
\label{eq:predict}
\end{align}
where $h_{\theta^n_s}(\tilde{s}^n)=\theta_s^n \tilde{s}^n + {b}^n$.
The function \(\rho\) signifies an appropriate transformation, such as the softmax function. It is worth noting that our model's use of element-wise summation ensures compliance with (\ref{eq:gate}) when operating on composite features. 
We denote the synthetic data as $\tilde{\mathcal{D}}$, while $\ty$ represents the conditional labels used in synthesizing $\tilde{{x}}$. The aggregate predictor's training is synchronized with the group selection models to update all the parameters $\Theta$ by minimizing a standard prediction loss (e.g., cross-entropy loss). 
The parameter \(\beta \ge 0\) offers a balance between the two terms. Unlike the conventional L1 regularization, we employ an L2 penalty on the averaged selection probability \(\bm\alpha^n\) for each individual learner. This choice is motivated by our aim to encourage smaller feature groups over larger ones.
    \begin{align}
        \label{eq:syn}
        \mathcal{L}_{\rm Syn} =  \mathbb{E}_{\tilde{\mathcal{D}}} \left[ \ell (\ty, \hat y ) \right] + \beta \sum_n^N\|\bm{\alpha}^n\|_2
    \end{align}

\end{enumerate}

Each individual selector should address the conventional feature selection problem, denoted by \(\mathcal{L}_{\rm Syn}\), which demands the group predictor to be precise while opting for minimal features. It is essential to note that while individual selectors are not inherently designed to be maximally predictive, overlapping features between groups are not strictly prohibited.

\subsection{Classification on Real Data}\label{sec:classification}
After updating the gate vectors using synthetic data, the same gate-filters are applied to select real data embeddings.

The steps are as follows:
\begin{enumerate}
    \item Gate Calculation: For real data, each gate \(\bm\alpha^n,\ n\in[N]\) is calculated using the learned gate parameters \(\textbf{w}^n\). These gates are then used to select the relevant embeddings:
    \begin{align}
    \alpha^n_i =  \mathbbm{1}\left({ w^n_i z_i>0}\right),  \quad  i \in [p].
    \label{eq:mask}
    \end{align}
    We obtain selected real embeddings from each selection gate $\cS_n$ via Eq.~\eqref{selection}.
    \item Prediction and Updating: The selected real embeddings are used to make predictions. The predictor parameters $\{\theta_s^n\}_{n=1}^N$ and the local encoder parameters $\{\theta_k\}_{k=1}^K$ are updated based on the real data. The selector parameters \(\{w^n\}_{n=1}^N\) are fixed during this stage. This trains the encoder to produce better representations of partial feature sets which in turn prevents the model selection from overfitting to spurious, noisy features, benefiting feature selection.
    We predict $\hat y$ via Eq. \eqref{eq:predict}, after that an overall loss is applied.
    \begin{align}
        \label{eq:real}
        \mathcal{L}_{\rm Real} =  \mathbb{E}_{\tilde{\mathcal{D}}} \left[ \ell (y, \hat y ) \right] + \beta \sum_n^N\|\bm{\alpha}^n\|_2
    \end{align}
\end{enumerate}

The objective
during this stage is to minimize the loss function \(\mathcal{L}_{\rm Real}\), ensuring that the ensemble network is a proficient predictor and that individual selectors solve the feature selection problem effectively.

The entire model is trained end-to-end using gradient descent, optimizing both the synthetic and real data objectives to ensure robust and privacy-preserving feature selection in VFL. The detailed procedure is summarized in Algorithm \ref{alg1}.

\begin{algorithm}[t]
\small
\caption{\small ICAFS Stage 2 \& 3: Inter-Client-Aware Feature Selection for Vertical Federated Learning}\label{ftl}
\begin{algorithmic}[1]
\State \textbf{Input: } 
Synthetic data $\tilde{\mathcal{D}}=(\{\tilde{X}_k\}_{k=1}^K, \tilde{Y})$ generated from pre-trained generative models $G_s, \{G_k\}_{k=1}^K$, and real data ${\mathcal{D}}=(\{{X}_k\}_{k=1}^K, {Y})$. ${X}_k, \tilde{X}_k \in \mathbb{R}^{d_k \times n_k}$ and $Y, \tilde{Y} \in \mathbb{R}^{n_k}$ are the concatenation of samples.
\State \textbf{Output: }  Optimal parameters $\Theta^*$  
\For{\(t \gets 1,\dots,T\)}
\State \textbf{Client} $k$ sends \(f_{\theta_k}(X_k)\), \(f_{\theta_k}(\tilde{X}_k)\), $\tilde{Y}$ to server. 
\State \textbf{Server:} \Comment{\textit{Selector update with synth data}}
\For{\(n \gets 1,\dots,N\) in parallel} 
\State Calculating $\bm\alpha^n$ based on \eqref{eq:alpha}.
\State Selecting synthetic embedding by \eqref{selection}. 
\State Obtaining prediction using \eqref{eq:predict}.
\State Updating $\Theta$ via minimizing \eqref{eq:syn}.
\EndFor
\State \textbf{Server:} \Comment{\textit{Classification with real data}}
\State Calculating gate $\bm\alpha^n$ again by \eqref{eq:mask}. 
\State Selecting real embedding by \eqref{selection}.
\State Obtaining prediction with \eqref{eq:predict} (replacing $\tilde{\textbf{s}}$ with $\textbf{s}$).
\State Updating $\{\theta_k\}_{k=1}^K$ and $\{\theta_s^n\}_{n=1}^N$ by minimizing \eqref{eq:real}.
\EndFor
\end{algorithmic}
\label{alg1}
\end{algorithm}

\section{Experiment}
This section is dedicated to assessing the efficacy of our proposed \ours{} across multi-modal and several biological and image datasets.

\subsection{Baselines}
The method under consideration is compared with four cutting-edge feature selection techniques, comprising three centralized methods and one VFL approach. The centralized methods were adapted to the VFL setting by implementing them within a federated framework where feature partitions are maintained across different parties while allowing necessary coordination for the selection process. The compared methods include:

\begin{itemize}
    \item SEFS \cite{lee2021}: Self-supervision enhanced feature selection with correlated gates 
    \item CompFS \cite{imrie2022}: Composite feature selection using deep ensembles 
    \item Group LASSO \cite{wang2021feature}: Feature selection using a neural network with group lasso regularization and controlled redundancy 
    \item VFLFS \cite{feng2022}: a method specifically designed for vertical federated learning
\end{itemize}

Group LASSO is chosen for comparison over LESS-VFL, as the authors \cite{castiglia2023} assert that LESS-VFL's performance is comparable to Group LASSO, and the code has not been made publicly available.

\subsection{Performance Metrics}
To evaluate the effectiveness of feature selection methods, we employ different metrics depending on the dataset type:

\begin{itemize}
    \item \textbf{Synthetic Experiments:} The true positive rate (TPR) is employed to measure the proportion of correctly identified relevant features compared to the known ground truth.
    
    \item \textbf{Real-world Datasets:} Since ground truth feature relevance is typically unknown in real-world data, we assess performance indirectly. We first apply each feature selection method to identify important features, then train a multi-layer perceptron (MLP) using only these selected features. The prediction performance of this MLP on test data serves as a proxy measure for the quality of the selected features.
\end{itemize}

\subsection{Implementation Details}
Our implementation includes the following components and configurations:

The gating mechanism employs a temperature parameter $\gamma$ ranging from 1 to 200, which controls the sharpness of the selection function used in determining whether a feature is relevant for a particular group. Higher values of $\gamma$ push the gating mechanism toward more binary decisions.

The encoder $f_\theta$ is implemented as a three-layer perceptron with ReLU activations, which transforms the input features into a representation suitable for the selection process.

For the Group Predictors $h_\theta$, we utilize a single linear layer for each group, implemented as a separate weight matrix and bias vector per group. This design allows each group to learn distinct patterns from the data.

The key hyperparameters for ICAFS are set as follows: $\beta=1.2$ and learning rate $LR=0.003$. 

\subsection{Description of Datasets}
We use four datasets to conduct our experiments as follows.
\begin{itemize}
\item  \noindent\textbf{Alzheimer’s Disease Neuroimaging Initiative (ADNI)} is a collaborative project aiming to enhance the clinical trials for Alzheimer’s disease (AD) prevention and treatment. It involves the collection, validation, and utilization of various data types, such as MRI and PET images, genetic information, cognitive assessments, and blood and CSF biomarkers, to predict the disease \cite{adni}. The ADNI dataset encompasses data from North American individuals of both genders, classified as ``Cognitively Normal,'' ``Mild Cognitive Impairment," or ``Alzheimer's Disease."

\item \noindent\textbf{TOX\_171} and \textbf{ALLAML} represent two biological datasets. As per \cite{ye2019distributed}, each of these datasets is randomly split equally along the feature dimension to simulate multi-party feature selection scenarios.

\item \noindent\textbf{USPS} is a dataset commonly utilized in fields such as computer vision and pattern recognition, particularly for handwritten text recognition \cite{usps}. It consists of 9,298 16×16 pixel grayscale images, each of which is centered and normalized, reflecting a wide variety of font styles.
\end{itemize}


\begin{table}
\small
\centering
\caption{\label{tab:dataset} \small Details of datasets in our experiment. ``\#samp'' means the number of samples, ``\#feat 1'' and ``\#feat 2'' refer to the number of features for modality 1 and 2 separately. In ADNI, modality 1 is SNPs and modality 2 is MRI.}
\begin{tabular}{l|rrrr}
\toprule
Dataset & \#samp  & \#feat 1  & \#feat 2 & \#class \\
\midrule
ADNI  & 1,470 & 3,518  & 328  & 3 \\
ALLAML & 72  & 7,129 & - & 2 \\
USPS & 9,298  & 256 & - & 10 \\
TOX\_171  & 171 & 5,748 & - & 4 \\
\bottomrule
\end{tabular}
\end{table}
A summary of the datasets is presented in Table \ref{tab:dataset}. For the ADNI dataset, which comprises two distinct modalities, SNPs (Single Nucleotide Polymorphisms, which are genetic variations where a single nucleotide in the genome differs between individuals) and MRI (Magnetic Resonance Imaging, which provides detailed brain structural images showing anatomy and potential atrophy patterns),
we have established two separate clients, each handling one modality. As for the other dataset, we have implemented a random split approach to allocate portions of the data to each client in VFL setup. Compared to baselines, which typically involve only two clients as a standard configuration in their experiments, our paper not only adheres to this default setting but also extends to accommodate a larger number of clients for enhanced comparative analysis in Table \ref{table multiclient}.

\subsection{Synthetic model}
Generators and discriminators are implemented using CNNs with 4 layers each. The generator performs a series of deconvolution operations to generate a record while a discriminator performs a series of convolution operations to classify a record. In each layer, a list of learnable filters ($3 \times 3$ matrices) is applied to the entire input matrix, i.e., convolution operations. Thus, the layer output size is proportional to the number of filters in each layer. For each pair of consecutive layers, it adds a Conv2d layer, a LayerNorm layer, and a LeakyReLU activation layer. The final layer is a Conv2d layer followed by a ReLU activation. 
\subsection{Architecture}
\label{implementation}
\ours~consists of four kinds of different network components: (i) local encoders $f_{\theta_i},\ i\in[K]$ (ii) feature selectors $S^n(\cdot;\bm\alpha^n),\ n\in[N]$, (iii) global encoder $f_{\theta_s}$, and (iv) predictors $h_\theta^n,\ n\in[N]$. We use a fully-connected network as the baseline architecture for network components (i)-(iv). Note that compared to traditional feature selection, the computational complexity/number of parameters scales linearly with the maximum permitted number of groups (in the same way that an ensemble scales with the number of members). Typically the number of groups will be relatively small, which should alleviate any practical concerns. The specific architectural details of \ours~are provided below:
\begin{itemize}
    \item Encoders $f_\theta$: Three-layer perceptron with ReLU activations, and two hidden layers of a given hidden width.
    \item Group Predictors $h_\theta$ : Single linear layer, given by a separate weight matrix and bias for each group.
\end{itemize}

\begin{table*}[t]
\caption{Accuracy comparison on benchmark datasets ($\text{mean}_\text{std}$). The best one in each column is in bold.}
\label{table multiclient}
\resizebox{\textwidth}{!}{
\begin{tabular}{l|ccc|ccc|ccc|c}
\toprule\multirow{2}{*}{Method}                                      & \multicolumn{3}{c|}{ALLAML}                 & \multicolumn{3}{c|}{USPS}                   & \multicolumn{3}{c|}{TOX\_171}  & \multicolumn{1}{c}{ADNI}              \\
                                                       & K=2          & K=5          & K=10         & K=2          & K=5          & K=10         & K=2           & K=5          & K=10  & K=2       \\ \midrule

VFLFS(2022)                                                 & \(98.85_{1.30}\) & \(98.04_{1.67}\) & \(91.44_{1.55}\) & \(68.77_{1.67}\) & \(58.97_{2.11}\) & \(46.24_{1.27}\) & \(74.65_{2.33}\)  & \(62.76_{2.55}\) & \(55.35_{1.78}\) &\(55.87_{0.76}\)\\
Group LASSO(2021)& \(98.76_{0.98}\) & \(98.21_{1.11}\) & \(92.67_{1.03}\) & \(70.45_{1.22}\) & \(59.97_{1.34}\) & \(45.44_{1.57}\) & \(75.455_{2.33}\) & \(60.21_{1.22}\) & \(55.46_{1.97}\) &\(55.37_{2.46}\)\\
CompFS(2022)                                                 & \(100_{0.00}\)    & \(99.32_{0.98}\) & \(91.57_{1.24}\) & \(70.51_{1.22}\) & \(55.15_{2.32}\) & \(44.32_{1.77}\) & \(73.01_{2.11}\)  & \(57.98_{1.56}\) & \(51.66_{1.66}\) &\(53.82_{1.78}\)\\
SEFS(2021)                                                   & \(97.56_{1.23}\)  & \(96.01_{0.78}\) & \(90.21_{1.14}\) & \(68.96_{2.11}\) & \(56.23_{1.78}\) & \(42.14_{2.56}\) & \(73.677_{2.78}\) & \(59.87_{2.06}\) & \(53.44_{2.54}\) &\(54.23_{1.26}\)\\ \midrule
ICAFS(ours)                                                  & \(\bm{100_{0.00}}\)   & \(\bm{100_{0.00}}\)   & \(\bm{93.33_{0.67}}\) & \(\bm{72.03_{1.25}}\) & \(\bm{60.23_{1.12}}\) & \(\bm{50.11_{1.44}}\) & \(\bm{76.28_{1.14}}\)  & \(\bm{63.84_{0.98}}\) & \(\bm{57.12_{1.57}}\) &\(\bm{57.00_{1.51}}\)\\ \bottomrule
\end{tabular}}
\end{table*}
\subsection{Performance Comparison and Scalability with Various Number of Clients} \label{sec:basic_compare}

We present the classification accuracy results of our proposed method and the comparison algorithms on multiple datasets in Table \ref{table multiclient}. Our proposed methods demonstrate superior or competitive performance across all datasets in the default setting with $K=2$.
While the baseline VFLFS exhibits strong performance, it does not match our methods' accuracy on dataset ALLAML. CompFS aligns with our gated approach on ALLAML but underperforms on USPS and TOX\_171, whereas SEFS shows the least consistency across evaluated datasets. A key advantage over CompFS is including a representation phase and considering inter-client correlations, enhancing cooperative classifier learning and highlighting embedding components' importance in feature selection. Performance nuances may stem from model strategies, training samples, and dataset characteristics.
Our study extends beyond the initial two-client scenarios, examining setups with more clients ($K=5,10$) for TOX\_171, ALLAML, and USPS. We observe decreasing accuracy on USPS as clients increase, but our approach exhibits only a slight reduction, demonstrating resilience and scalability by benefiting from reduced feature synthesis complexity when features per client decrease with more clients. This resilience proves valuable in fragmented data environments, indicating robust scalability across client configurations. Generally, our approach surpasses compared algorithms under different settings, confirming its efficacy in feature selection.


\subsection{Performance on Multi-modal Data - ADNI}
As delineated in Section \ref{sec:classification}, the act of concatenating embedding components from all participating parties is anticipated to enhance the efficacy of feature selection. 
To substantiate the merit of intermediate representation concatenation, a comparative analysis of classification accuracy between \textit{\ours} and the baseline methods is conducted on the real-world multi-modal dataset ADNI, with one client with MRI data and another with SNPs data. Compared to the experiments in Section \ref{sec:basic_compare}, the multi-modal data consists of a wider range of different features that have underlying correlations. The outcomes, as shown in the last column of Table \ref{table multiclient}, consistently favor \textit{\ours}.
This improvement can be traced back to primary factors that \ours{} effectively explores the hidden interaction between modalities, mitigating the information shortfall at individual parties by facilitating information interchange during the training phase.

\subsection{Tradeoff between privacy protection and accuracy}
\label{sec:dp}

\begin{table}
\centering
\caption{Performance with DP-SGD ($\text{mean}_\text{std}$).}
\label{dp-sgd}
\resizebox{0.95\columnwidth}{!}{
\begin{tabular}{l|llll}
\toprule
\((\epsilon,\delta)\)                             & ADNI  & ALLAML        & USPS         & TOX\_171     \\ \midrule
N/A                          &    \(57.00_{1.51}\)   & \(100.00_{0.00}\) & \(72.03_{1.25}\) & \(76.28_{1.14}\) \\
$(10,1e^{-5})$               & \(55.96_{1.76}\)& \(93.33_{0.14}\)& \(71.20_{1.13}\)&\(57.14_{1.32}\) \\
$(5,1e^{-5})$                    & \(54.97_{1.37}\) & \(93.33_{0.22}\) & \(63.87_{0.35}\) & \(54.29_{1.17}\) \\ \bottomrule
\end{tabular}}
\end{table}

To investigate the trade-off between privacy protection and computational efficiency, we used Differential Privacy Stochastic Gradient Descent (DP-SGD) as a supplementary analysis. DP-SGD quantifies privacy guarantees by bounding the impact of an individual's data on the analysis outcome \cite{abadi2016deep} through the parameters $(\epsilon, \delta)$, where $\epsilon$ limits the change in output probabilities, and $\delta$ allows for a small chance that this bound is exceeded - lower values preserve privacy better. 

Our method maintained competitive performance even with DP-SGD introduced, as shown in Table \ref{dp-sgd}. The table presents the mean classification accuracy (\%) with standard deviations as subscripts across four datasets under different privacy settings. The baseline performance without differential privacy (N/A) achieves the highest accuracy across all datasets (e.g., 100\% on ALLAML, 72.03\% on USPS). As privacy constraints strengthen (smaller $\epsilon$ values), we observe a gradual performance decline, with the most significant drops occurring in the TOX\_171 dataset (from 76.28\% to 54.29\% at $\epsilon=5$), while ADNI shows better resilience (only 2.03\% decrease at $\epsilon=5$). Despite these promising results, implementing DP-SGD substantially increased computational overhead, which is why we did not ultimately adapt it in our final approach. Nevertheless, this analysis provides valuable insights into the quantifiable privacy-utility tradeoff, demonstrating that moderate privacy guarantees ($\epsilon=10$) can be achieved with relatively small performance sacrifices (average 
6.92\% accuracy reduction across datasets vs 9.71\% when $\epsilon=5$) compared to non-private alternatives.

\subsection{Learnable Gate Analysis}

\begin{table}
\caption{Ablation studies on introduce of learnable gate ($\text{mean}_\text{std}$).}
\label{framework comparison}
\resizebox{\columnwidth}{!}{
\begin{tabular}{l|llll}
\toprule
Accuracy                             & ADNI  & ALLAML        & USPS         & TOX\_171     \\ \midrule
\ours                          &    {\(\bm{57.00_{1.51}}\)}   & \(\bm{100.00_{0.00}}\) & \(\bm{72.03_{1.25}}\) & \(\bm{76.28_{1.14}}\) \\
\ours~(fixed temp) &\(56.53_{1.23}\)& \(100.00_{0.00}\)& \(71.25_{1.03}\)& \(73.13_{1.11}\) \\
\ours~(embedding)                     & \(55.76_{0.37}\) & \(100.00_{0.00}\) & \(69.25_{1.32}\) & \(72.31_{1.91}\) \\
W\textbackslash{}O feature selection & \(52.489_{2.33}\)   &  \(98.76_{0.98}\)  & \(65.16_{1.23}\)  &  \(65.44_{2.33}\)      \\ \bottomrule
\end{tabular}}
\end{table}
We also conduct an ablation study on the FS mechanism. In particular, we made three modifications to our original \ours{}.
\begin{itemize}
    \item \ours~(fixed temp): We fix the temperature in (\ref{eq:alpha}) as 1
    \item \ours~(embedding): We simply use an MLP for feature selection on the original embedding
    \item W\textbackslash{}O feature selection: We remove the feature selection module 
\end{itemize}  
The comparison demonstrates the advantage of learnable gate mechanism as shown in Table \ref{framework comparison}, achieving superior accuracy across various datasets including ADNI, ALLAML, USPS, and TOX\_171. Furthermore, the significant improvement with FS highlights the importance of our FS mechanism. 
\subsection{Qualitative Assessment of Synthetic Features}

\begin{figure*}
    \centering
    \begin{minipage}{0.37\columnwidth}
        \includegraphics[width=1\linewidth]{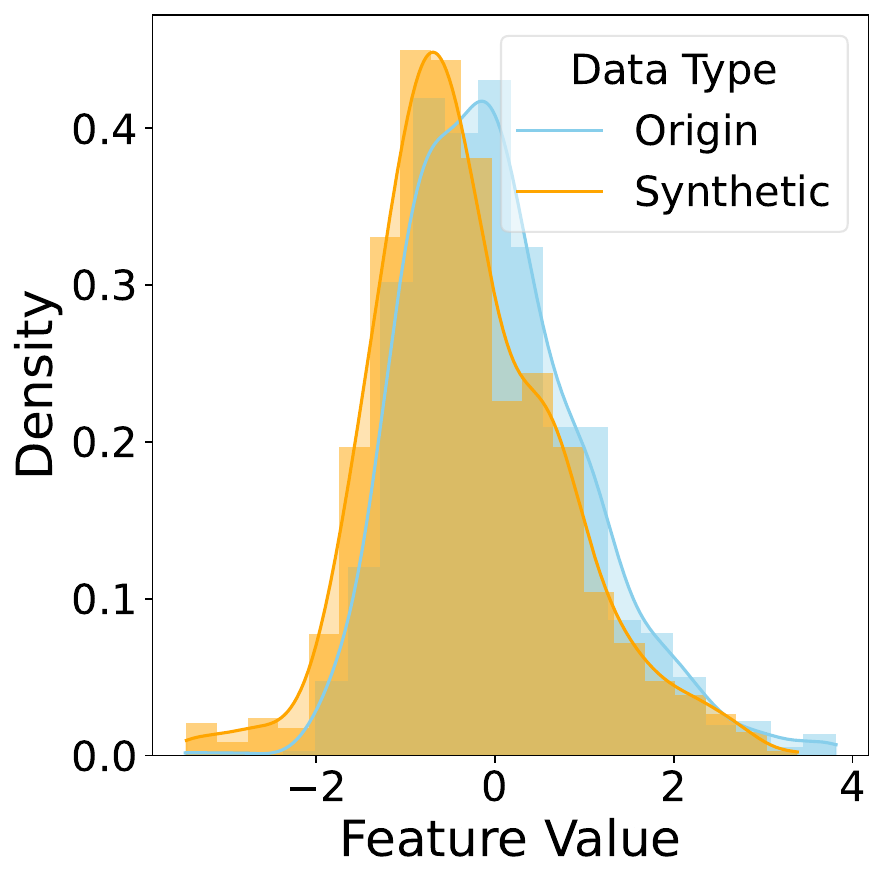}
        \centerline{\small (a) CallosumMidPosterior}
    \end{minipage}
    \hspace{0.1cm}
    \begin{minipage}{0.37\columnwidth}
        \includegraphics[width=1\linewidth]{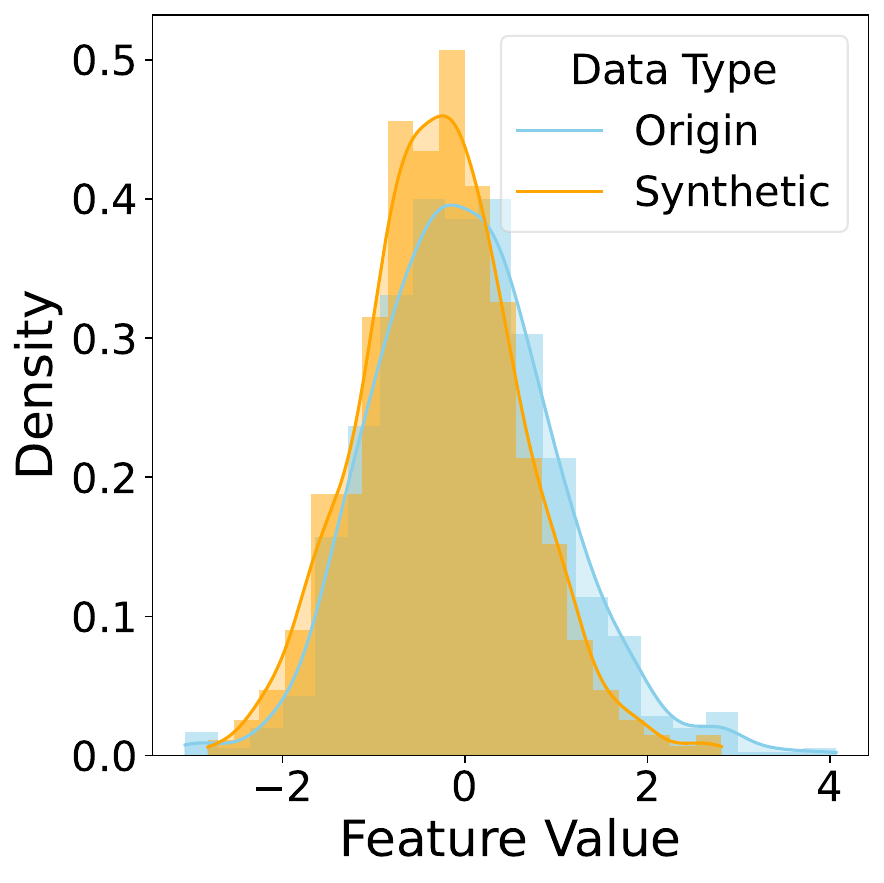}
        \centerline{\small (b) CorpusCallosumAnterior}
    \end{minipage}
    \begin{minipage}{0.395\columnwidth}
        \includegraphics[width=1\linewidth]{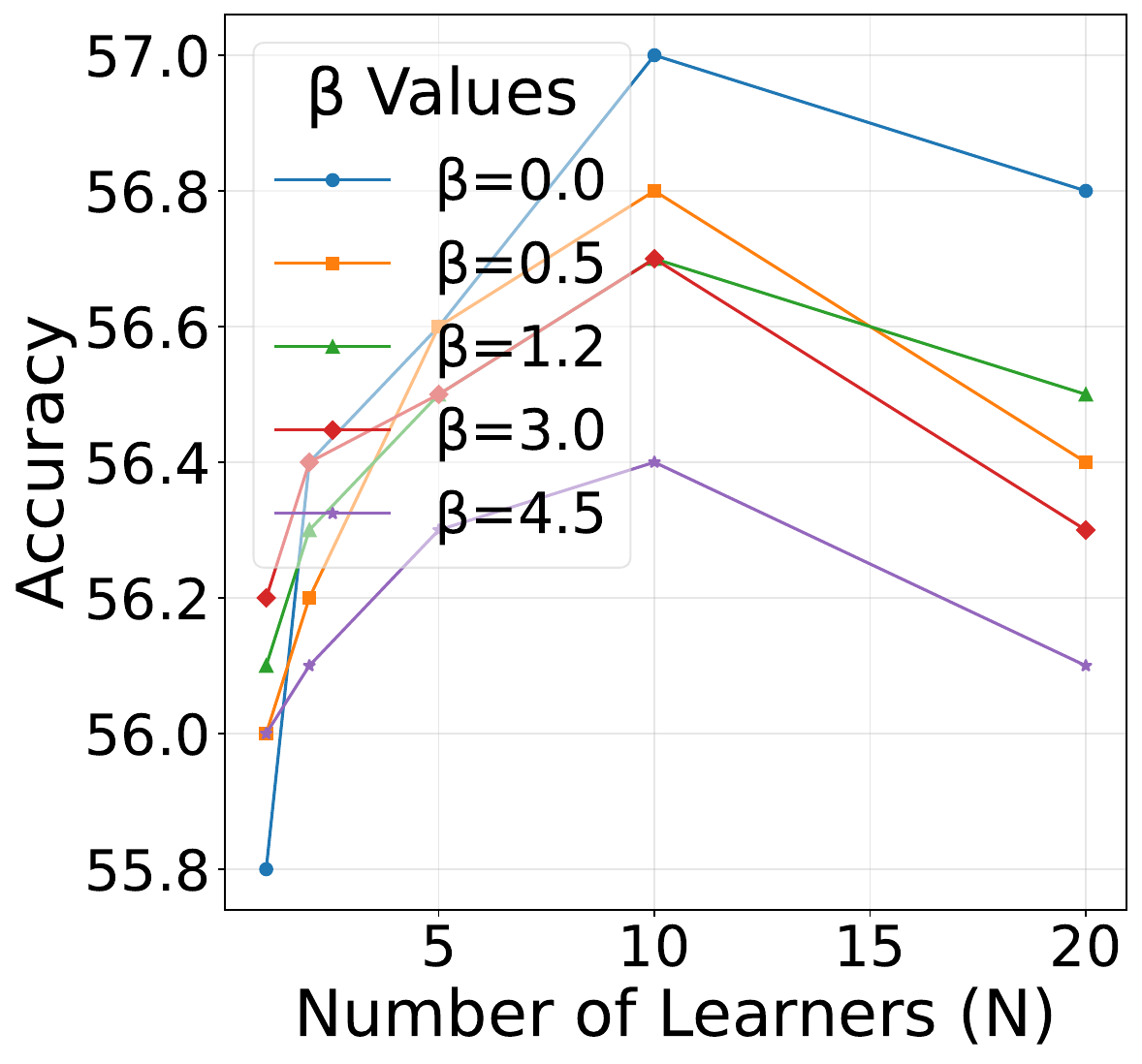}
        \centerline{\small (c) ADNI}
    \end{minipage}
    \begin{minipage}{0.395\columnwidth}
        \includegraphics[width=1\linewidth]{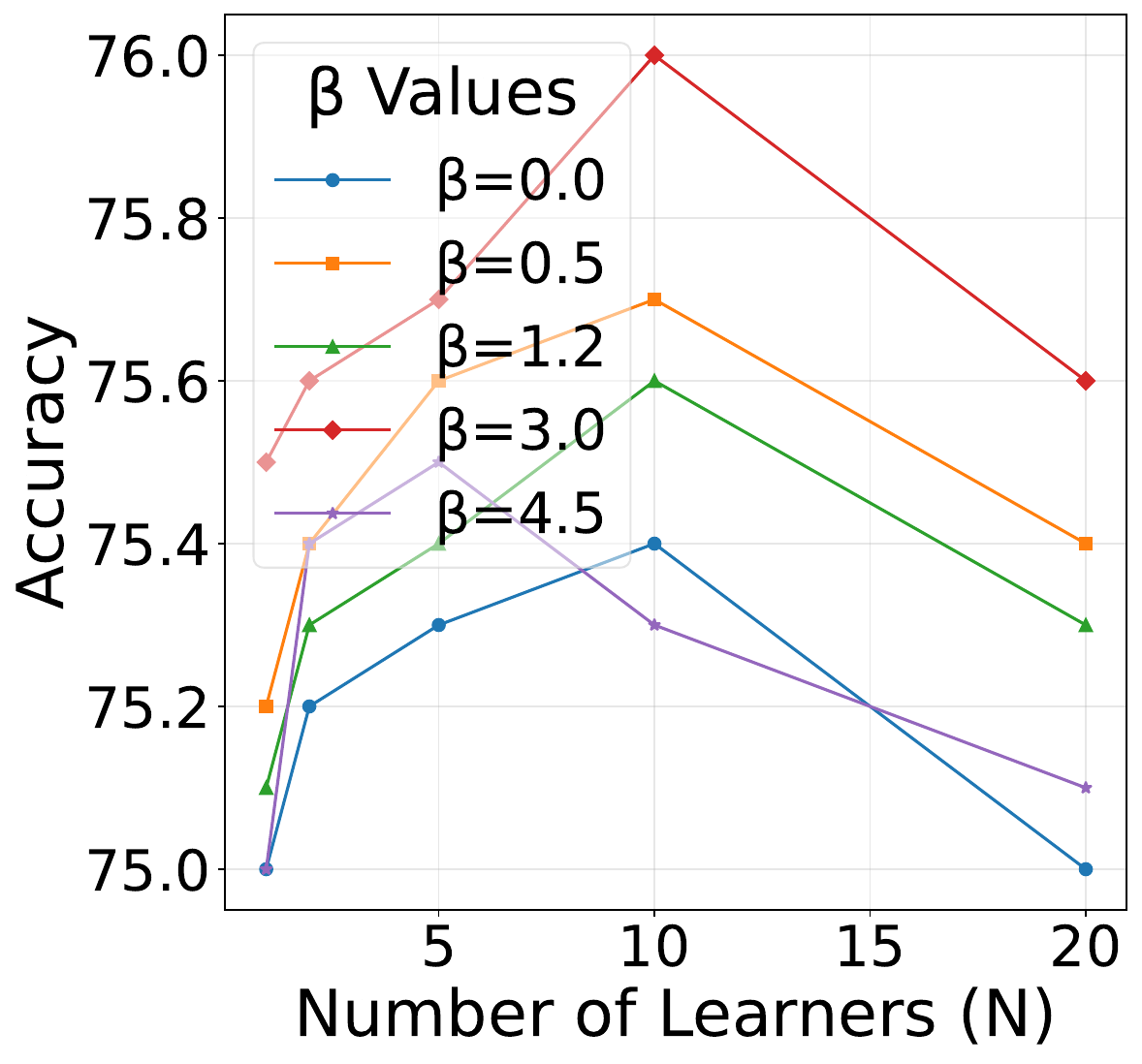}
        \centerline{\small (d) Tox}
    \end{minipage}
    \begin{minipage}{0.37\columnwidth}
        \includegraphics[width=1\linewidth]{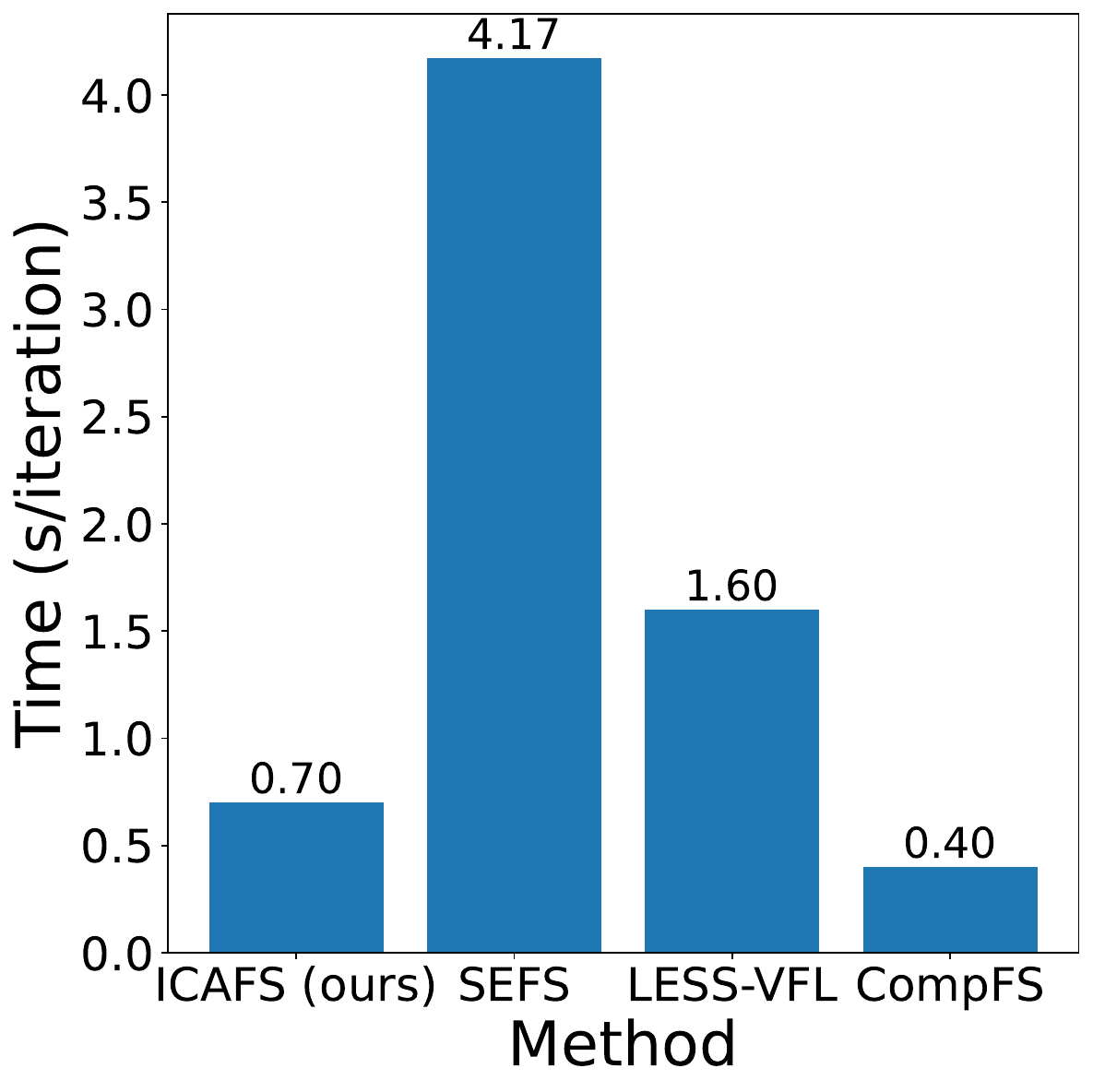}
        \centerline{\small (e) Training time comparison}
    \end{minipage}
    \caption{(a-b) Distribution of both original and synthetic data. (c-d) Ablation investigating on $N$ and $\beta$. (e) Training time comparison.}
    \label{fig:distribution}
\end{figure*}


We conducted an empirical analysis using the ADNI dataset as a reference to quantitatively visualize the fitness of our synthetic features. Specifically, we selected two features (Corpus Callosum Anterior Vanishing White Matter Disease and Corpus Callosum Mid Posterior Vanishing White Matter Disease), which are significant factors for Alzheimer’s Disease. The histograms of the feature values of both the original and the synthetic data are presented in Fig. \ref{fig:distribution}~(a-b)
that demonstrates a close alignment between the two distributions, indicating that our data synthesis model has a remarkable capacity for capturing the complexity and variability inherent in real-world data distributions. 
\subsection{Ablation investigating on Client Number $N$ and Regularization $\beta$.} 
To explore the behavior of \ours~further, we conducted a sensitivity analysis of several hyperparameters on ADNI and TOX dataset. The results are shown in Figure~\ref{fig:distribution}~(c-d).
Increasing the number of selectors generally leads to an increase in accuracy. This is evident from the progression from $N=1$ to $N=20$, where there is a slight upward trend in accuracy across most $\beta$ values. However, this increase is not linear and shows diminishing returns, especially beyond $N=5$.
The value of $\beta$ also influences accuracy, with mid-range $\beta$ values ($\beta=1.2$) generally resulting in the highest accuracy across different selector counts. This indicates that there is a balance in the strength of regularization (as $\beta$ might be interpreted in some models) to optimize performance.
Thus, $N=5$ with $\beta=1.2$ might be the sweet spot, offering high accuracy with relatively lower variability. It seems to offer a good trade-off between performance and stability without requiring as many resources as higher selector counts.


\subsection{Training Efficiency}
\label{app:efficiency}

Our training methodology incorporates an additional stage of federated synthesis. While this stage adds some computational overhead during the synthesis phase, the overall approach remains more efficient than the complete training process with synthetic features. This efficiency is clearly illustrated in Table~\ref{tab:cost}, which presents an evaluation using the ADNI dataset. It is important to note that when compared to simpler baseline methods such as CompFS, VFLFS, and Group LASSO, our approach maintains comparable computational costs for both the training with synthetic features and the inference time. Although our method requires a modest additional cost for the synthesis phase, it remains computationally efficient, especially when compared with more complex baselines like SEFS, where we observe a substantial reduction in overall costs. This cost efficiency, despite the additional federated synthesis stage, demonstrates the practicality of our approach in scenarios where computational resources and time are limited.

As shown in Figure \ref{fig:distribution}~(e), we can see that the time per iteration in Stage 2 and Stage 3
of our method outperforms SEFS and LESS-VFL and is competitive compared with CompFS. Note that the accuracy performance of our method is significantly better than CompFS. Besides, the synthetic data generation process costs 7.7s per iteration, which seems longer than the training stage. However, the number of iterations is small, and the total time consumption of the synthetic data generation stage is similar to the training stage. Furthermore, the synthetic data generation (Stage 1) is a completely separate process from the training stage and can be conducted in advance, which is not a concern in terms of time complexity. To sum up, our method is pretty efficient and outperforms most state-of-the-art baseline methods in terms of time complexity.

\begin{table}
\small
\setlength{\tabcolsep}{2pt}
\centering
\caption{\small Computational (comp) and communication (comm) costs.}
\label{tab:cost}
\begin{tabular}{c|cc|cc}
\hline
& \multicolumn{2}{c|}{Synthetic Stage} & \multicolumn{2}{c}{Training Stage} \\
& Comm (mb) & Comp (gmac) & Comm (mb) & Comp (gmac) \\ 
\hline
SEFS   & /   & /    & 14018.7M   & 109672.2K \\
ICAFS   & 967.93M   &  7657.4K  & 2300.1M   & 18034.5K \\ 
\hline
\end{tabular}
\end{table}



\begin{table}
\small
\centering
\caption{Robustness Validation on Gaussian Noise.}
\label{robustness}
\resizebox{\columnwidth}{!}{
\begin{tabular}{lllll}
\hline
Accuracy(\%) & ADNI   & ALLAML & USPS   & TOX\_171 \\ \hline
w/o noise    &    \(57.00_{1.51}\)   & \(100.00_{0.00}\) & \(72.03_{1.25}\) & \(76.28_{1.14}\) \\ \hline
Noise 50\%   & \(55.081_{1.32}\) & \(100_{0.00}\)    & \(50.696_{2.54}\) & \(57.143_{1.79}\)   \\
Noise 33\%   & \(55.018_{1.91}\) & \(93.333_{1.03}\) & \(57.246_{3.23}\) & \(62.57_{2.58}\)    \\
Noise 20\%   & \(54.872_{0.98}\) & \(100_{0.00}\)    & \(65.766_{2.42}\) & \(65.244_{3.03}\)   \\ \hline
\end{tabular}}
\end{table}
\subsection{Robustness Validation on Privacy-preserving Random Noise} \label{app:noise}
Besides the proven efficacy of \textit{\ours}, a robustness assessment using Gaussian noise further underscores the durability of the proposed method. This evaluation involved the injection of different noise levels (50\%, 33\%, and 20\%) into the ADNI, ALLAML, USPS, and TOX\_171 datasets as shown in Table \ref{robustness}. These artificial features serve as non-significant attributes, enabling an assessment of the feature selection capability. Consequently, the ultimate test accuracy stands as a testament to the correct identification of significant features and the training of a well-generalizing model. Remarkably, even in the face of considerable noise, our approach sustained acceptable accuracy across all datasets. The performance on the ALLAML dataset is particularly striking, achieving 100\% accuracy despite 20\% noise. This resilience to noise is an essential quality, especially in real-world applications where data might be exposed to various noise types and alterations. These findings endorse our method as a trustworthy option for vertical federated learning, adept at navigating the complexities of noisy conditions.
\subsection{Further Discussion}
\textbf{Communication Overhead.}
Although \ours~requires additional synthetic data generation, the process happens before the FL training, and can be preprocessed offline. We discuss the efficiency of \ours~in Section \ref{app:efficiency} compared to standard VFL baselines and show that \ours~has comparable or even better communication overhead.

\textbf{Privacy.} 
The purpose of introducing synthetic data for feature selection in \ours~is to reduce data information leakage. Previous work empirically showed that synthetic data can protect against some privacy attacks \cite{huang2024overcoming}, such as member inference attacks \cite{shokri2017membership} and gradient inversion attacks \cite{huang2021evaluating}. In the generation of synthetic data, we only shared gradient-level information. The existing data inference methods for data distribution to recover the exact training data typically require obtaining instance-level information from training data \cite{zhang2020secret}, which is not available in our setting. 
\section{conclusion}
This paper introduced a new framework for feature selection in VFL, designed to facilitate cooperative inter-client feature selection without exposing local data and attaining unparalleled performance in applicable VFL scenarios. We novelly introduce label-conditional synthetic data to facilitate FS with a learnable gating mechanism. 
Through rigorous experimentation on various real-world datasets, we demonstrated that our model is capable of identifying features that yield enhanced predictive accuracy. Unlike many feature selection approaches, the proposed framework is a universal solution, imposing no constraints on the nature or category of the data. 

In the future, taking into account the triple V framework ~ \cite{laney20013d} during this big data era—characterized by rapid velocity, substantial volume, and wide variety—we might think about modifying our approach for online learning applications~\cite{hou2017learning,hou2021prediction,hou2021storage,he2021online} to enhance practicality and integrating blockchain framework to enhance security~\cite{ruochen}.

\section*{acknowledge}
The authors would like to thank Xiaoxiao Li, for her support in this manuscript. This work was supported in part by the NIH grants U01 AG068057 and U01 AG066833. Data collection and sharing for this project was funded by the Alzheimer’s Disease Neuroimaging Initiative (ADNI) (National Institutes of Health Grant U01 AG024904) and DOD ADNI (Department of Defense award number W81XWH-12-2-0012). ADNI is funded by the National Institute on Aging, the National Institute of Biomedical Imaging and Bioengineering, and through generous contributions from the following: AbbVie, Alzheimer’s Association; Alzheimer’s Drug Discovery Foundation; Araclon Biotech; BioClinica, Inc.; Biogen; Bristol-Myers Squibb Company; CereSpir, Inc.; Cogstate; Eisai Inc.; Elan Pharmaceuticals, Inc.; Eli Lilly and Company; EuroImmun; F. Hoffmann-La Roche Ltd and its affiliated company Genentech, Inc.; Fujirebio; GE Healthcare; IXICO Ltd.; Janssen Alzheimer Immunotherapy Research \& Development, LLC.; Johnson \& Johnson Pharmaceutical Research \& Development LLC.; Lumosity; Lundbeck; Merck \& Co., Inc.; Meso Scale Diagnostics, LLC.; NeuroRx Research; Neurotrack Technologies; Novartis Pharmaceuticals Corporation; Pfizer Inc.; Piramal Imaging; Servier; Takeda Pharmaceutical Company; and Transition Therapeutics. The Canadian Institutes of Health Research is providing funds to support ADNI clinical sites in Canada. Private sector contributions are facilitated by the Foundation for the National Institutes of Health (www.fnih.org). The grantee organization is the Northern California Institute for Research and Education, and the study is coordinated by the Alzheimer’s Therapeutic Research Institute at the University of Southern California. ADNI data are disseminated by the Laboratory for Neuro Imaging at the University of Southern California.

\balance
\bibliography{icafs}

\begin{thebibliography}{10}

\bibitem{li2020review}
L.~Li, Y.~Fan, M.~Tse, and K.-Y. Lin, ``A review of applications in federated learning,'' {\em Computers \& Industrial Engineering}, vol.~149, p.~106854, 2020.

\bibitem{qin2021federated}
Y.~Qin and M.~Kondo, ``Federated learning-based network intrusion detection with a feature selection approach,'' in {\em 2021 International Conference on Electrical, Communication, and Computer Engineering (ICECCE)}, pp.~1--6, IEEE, 2021.

\bibitem{ahmed2021federated}
U.~Ahmed, G.~Srivastava, and J.~C.-W. Lin, ``A federated learning approach to frequent itemset mining in cyber-physical systems,'' {\em Journal of Network and Systems Management}, vol.~29, pp.~1--17, 2021.

\bibitem{chen2020}
T.~Chen, X.~Jin, Y.~Sun, and W.~Yin, ``Vafl: A method of vertical asynchronous federated learning,'' {\em arXiv preprint arXiv:2007.06081}, 2020.

\bibitem{jackson2018genetic}
M.~Jackson, L.~Marks, G.~H. May, and J.~B. Wilson, ``The genetic basis of disease,'' {\em Essays in biochemistry}, vol.~62, no.~5, pp.~643--723, 2018.

\bibitem{Wu2023FederatedAL}
X.~Wu, J.~Pei, C.~Chen, {\em et~al.}, ``Federated active learning for multicenter collaborative disease diagnosis,'' {\em IEEE Transactions on Medical Imaging}, vol.~42, no.~7, pp.~2068--2080, 2023.

\bibitem{jiang2017high}
B.~Jiang, X.~Wang, and C.~Leng, ``High-dimensional biological feature selection by sparse group lasso with oracle property,'' in {\em Proceedings of the 34th International Conference on Machine Learning}, pp.~1748--1756, PMLR, 2017.

\bibitem{li2015feature}
J.~Li, K.~Cheng, S.~Wang, F.~Morstatter, R.~P. Trevino, J.~Tang, and H.~Liu, ``Feature selection for high-dimensional genomic data,'' in {\em Proceedings of the 21th ACM SIGKDD International Conference on Knowledge Discovery and Data Mining}, pp.~1215--1224, 2015.

\bibitem{zhang2018feature}
Y.~Zhang, C.~Bernau, G.~Parmigiani, and L.~Waldron, ``Feature selection in high-dimensional precision medicine data,'' in {\em Proceedings of the 2018 ACM International Conference on Bioinformatics, Computational Biology, and Health Informatics}, pp.~91--100, 2018.

\bibitem{sun2019high}
W.~Sun, P.~Wang, D.~Yin, J.~Yang, and Y.~Chang, ``High-dimensional feature selection for sample efficient treatment effect estimation,'' in {\em Advances in Neural Information Processing Systems}, pp.~11673--11683, 2019.

\bibitem{chen2017kernel}
J.~Chen, M.~Stern, M.~J. Wainwright, and M.~I. Jordan, ``Kernel feature selection via conditional covariance minimization,'' in {\em NeurIPS}, 2017.

\bibitem{song2012feature}
L.~Song, A.~Smola, A.~Gretton, J.~Bedo, and K.~Borgwardt, ``Feature selection via dependence maximization,'' {\em Journal of Machine Learning Research}, vol.~13, no.~5, 2012.

\bibitem{estevez2009normalized}
P.~A. Est\'evez, M.~Tesmer, C.~A. Perez, and J.~M. Zurada, ``Normalized mutual information feature selection,'' {\em IEEE Transactions on neural networks}, vol.~20, no.~2, pp.~189--201, 2009.

\bibitem{roy2015feature}
D.~Roy, K.~S.~R. Murty, and C.~K. Mohan, ``Feature selection using deep neural networks,'' in {\em International Joint Conference on Neural Networks (IJCNN)}, pp.~1--6, IEEE, 2015.

\bibitem{kabir2010new}
M.~M. Kabir, M.~M. Islam, and K.~Murase, ``A new wrapper feature selection approach using neural network,'' {\em Neurocomputing}, vol.~73, no.~16-18, pp.~3273--3283, 2010.

\bibitem{yamada2020feature}
Y.~Yamada, O.~Lindenbaum, S.~Negahban, and Y.~Kluger, ``Feature selection using stochastic gates,'' in {\em International Conference on Machine Learning}, pp.~10648--10659, PMLR, 2020.

\bibitem{li2016deep}
Y.~Li, C.-Y. Chen, and W.~W. Wasserman, ``Deep feature selection: theory and application to identify enhancers and promoters,'' {\em Journal of Computational Biology}, vol.~23, no.~5, pp.~322--336, 2016.

\bibitem{hans2009bayesian}
C.~Hans, ``Bayesian lasso regression,'' {\em Biometrika}, vol.~96, no.~4, pp.~835--845, 2009.

\bibitem{feng2020multi}
S.~Feng and H.~Yu, ``Multi-participant multi-class vertical federated learning,'' {\em arXiv preprint arXiv:2001.11154}, 2020.

\bibitem{chen2021fedeini}
X.~Chen, S.~Zhou, B.~Guan, K.~Yang, H.~Fao, H.~Wang, and Y.~Wang, ``Fed-eini: An efficient and interpretable inference framework for decision tree ensembles in vertical federated learning,'' in {\em IEEE Int. Conf. Big Data}, pp.~1242--1248, 2021.

\bibitem{hou2022}
J.~Hou, M.~Su, A.~Fu, and Y.~Yu, ``Verifiable privacy-preserving scheme based on vertical federated random forest,'' {\em IEEE Internet Things}, vol.~9, no.~22, pp.~22158--22172, 2022.

\bibitem{zhang2022a}
Y.~Zhang, Y.~Hu, X.~Gao, D.~Gong, Y.~Guo, K.~Gao, and W.~Zhang, ``An embedded vertical-federated feature selection algorithm based on particle swarm optimisation,'' {\em CAAI Trans. Intel. Techn.}, 2022.

\bibitem{zhang2022b}
R.~Zhang, H.~Li, M.~Hao, H.~Chen, and Y.~Zhang, ``Secure feature selection for vertical federated learning in ehealth systems,'' in {\em IEEE Int. Conf. Comm.}, pp.~1257--1262, 2022.

\bibitem{chen2022}
P.~Chen, X.~Du, Z.~Lu, J.~Wu, and P.~C.~K. Hung, ``Evfl: an explainable vertical federated learning for data-oriented artificial intelligence systems,'' {\em J. Syst. Archit.}, vol.~126, p.~102474, 2022.

\bibitem{li2023}
A.~Li, H.~Peng, L.~Zhang, J.~Huang, Q.~Guo, H.~Yu, and Y.~Liu, ``Fedsdg-fs: Efficient and secure feature selection for vertical federated learning,'' in {\em IEEE Int. Conf. Comp. Comm.}, 2023.

\bibitem{feng2022}
S.~Feng, ``Vertical federated learning-based feature selection with non-overlapping sample utilization,'' {\em Expert Systems with Applications}, vol.~208, p.~118097, 2022.

\bibitem{castiglia2023}
T.~Castiglia, Y.~Zhou, S.~Wang, S.~R. Kadhe, N.~B. Angel, and S.~Patterson, ``Less-vfl: Communication-efficient feature selection for vertical federated learning,'' in {\em International Conference on Machine Learning}, July 2023.

\bibitem{fu2022label}
C.~Fu, X.~Zhang, S.~Ji, J.~Chen, J.~Wu, S.~Guo, J.~Zhou, A.~X. Liu, and T.~Wang, ``Label inference attacks against vertical federated learning,'' in {\em 31st USENIX Security Symposium (USENIX Security 22)}, pp.~1397--1414, 2022.

\bibitem{liu2022batch}
Y.~Liu, T.~Zou, Y.~Kang, W.~Liu, Y.~He, Z.~Yi, and Q.~Yang, ``Batch label inference and replacement attacks in black-boxed vertical federated learning,'' 2022.

\bibitem{vu2023}
M.~N. Vu, T.~R. Jeter, R.~Alharbi, and M.~T. Thai, ``Active data reconstruction attacks in vertical federated learning,'' in {\em 2023 IEEE International Conference on Big Data (BigData)}, pp.~1374--1379, 2023.

\bibitem{zhao2023}
Z.~Zhao, H.~Wu, A.~V. Moorsel, and L.~Y. Chen, ``Gtv: Generating tabular data via vertical federated learning,'' {\em arXiv preprint arXiv: 2302.01706}, 2023.

\bibitem{lee2021}
J.~Lee and M.~van~der Schaar, ``Self-supervision enhanced feature selection with application to medical diagnosis,'' in {\em International Conference on Machine Learning}, pp.~6144--6153, 2021.

\bibitem{imrie2022}
F.~Imrie, A.~Norcliffe, P.~Liò, and M.~van~der Schaar, ``Composite feature selection using deep ensembles,'' {\em Advances in Neural Information Processing Systems}, vol.~35, pp.~36142--36160, 2022.

\bibitem{wu2022}
Y.~Wu, Y.~Kang, J.~Luo, Y.~He, L.~Fan, R.~Pan, and Q.~Yang, ``Fedcg: Leverage conditional gan for protecting privacy and maintaining competitive performance in federated learning,'' International Joint Conferences on Artificial Intelligence Organization, jul 2022.

\bibitem{amaldi1998approximability}
E.~Amaldi and V.~Kann, ``On the approximability of minimizing nonzero variables or unsatisfied relations in linear systems,'' {\em Theoretical Computer Science}, vol.~209, no.~1-2, pp.~237--260, 1998.

\bibitem{lyu2023}
F.~Lyu, X.~Tang, D.~Liu, L.~Chen, X.~He, and X.~Liu, ``Optimizing feature set for click-through rate prediction,'' in {\em Proceedings of the {ACM} Web Conference 2023, {WWW} 2023}, (Austin, TX, USA), pp.~3386--3395, {ACM}, 2023.

\bibitem{wang2021feature}
J.~Wang, H.~Zhang, J.~Wang, Y.~Pu, and N.~R. Pal, ``Feature selection using a neural network with group lasso regularization and controlled redundancy,'' {\em IEEE Transactions on Neural Networks and Learning Systems}, vol.~32, no.~3, pp.~1110--1123, 2021.

\bibitem{adni}
A.~D.~N. Initiative, ``Adni,'' 2004.

\bibitem{ye2019distributed}
X.~Ye, H.~Li, A.~Imakura, and T.~Sakurai, ``Distributed collaborative feature selection based on intermediate representation.,'' in {\em IJCAI}, pp.~4142--4149, 2019.

\bibitem{usps}
U.~P. Service, ``Usps,'' 1998.

\bibitem{abadi2016deep}
M.~Abadi, A.~Chu, I.~Goodfellow, H.~B. McMahan, I.~Mironov, K.~Talwar, and L.~Zhang, ``Deep learning with differential privacy,'' in {\em Proceedings of the 2016 ACM SIGSAC Conference on Computer and Communications Security}, pp.~308--318, ACM, 2016.

\bibitem{huang2024overcoming}
C.-Y. Huang, K.~Srinivas, X.~Zhang, and X.~Li, ``Overcoming data and model heterogeneities in decentralized federated learning via synthetic anchors,'' 2024.

\bibitem{shokri2017membership}
R.~Shokri, M.~Stronati, C.~Song, and V.~Shmatikov, ``Membership inference attacks against machine learning models,'' in {\em 2017 IEEE Symposium on Security and Privacy (SP)}, pp.~3--18, IEEE, 2017.

\bibitem{huang2021evaluating}
Y.~Huang, S.~Gupta, Z.~Song, K.~Li, and S.~Arora, ``Evaluating gradient inversion attacks and defenses in federated learning,'' {\em Advances in Neural Information Processing Systems}, vol.~34, pp.~7232--7241, 2021.

\bibitem{zhang2020secret}
Y.~Zhang, R.~Jia, H.~Pei, W.~Wang, B.~Li, and D.~Song, ``The secret revealer: Generative model-inversion attacks against deep neural networks,'' in {\em Proceedings of the IEEE/CVF Conference on Computer Vision and Pattern Recognition}, pp.~253--261, 2020.

\bibitem{laney20013d}
D.~Laney, ``3d data management: Controlling data volume, velocity, and variety,'' tech. rep., META Group, February 2001.
\newblock META Group Research Note.

\bibitem{hou2017learning}
B.-J. Hou, L.~Zhang, and Z.-H. Zhou, ``Learning with feature evolvable streams,'' {\em Advances in Neural Information Processing Systems}, vol.~30, 2017.

\bibitem{hou2021prediction}
B.-J. Hou, L.~Zhang, and Z.-H. Zhou, ``Prediction with unpredictable feature evolution,'' {\em IEEE Transactions on Neural Networks and Learning Systems}, vol.~33, no.~10, pp.~5706--5715, 2021.

\bibitem{hou2021storage}
B.-J. Hou, Y.-H. Yan, P.~Zhao, and Z.-H. Zhou, ``Storage fit learning with feature evolvable streams,'' in {\em Proceedings of the AAAI Conference on Artificial Intelligence}, vol.~35, pp.~7729--7736, 2021.

\bibitem{he2021online}
Y.~He, J.~Dong, B.-J. Hou, Y.~Wang, and F.~Wang, ``Online learning in variable feature spaces with mixed data,'' in {\em 2021 IEEE International Conference on Data Mining (ICDM)}, pp.~181--190, IEEE, 2021.

\bibitem{ruochen}
R.~Jin, B.~Wei, Y.~Luo, T.~Ren, and R.~Wu, ``Blockchain-based data collection with efficient anomaly detection for estimating battery state-of-health,'' {\em IEEE Sensors Journal}, vol.~21, no.~12, pp.~13455--13465, 2021.

\end{thebibliography}
\bibliographystyle{ieeetr}
\end{document}